\documentclass[runningheads]{llncs}

\usepackage{graphicx}
\usepackage{comment}
\usepackage{amsmath,amssymb}
\usepackage{color}
\usepackage{url}
\usepackage{hyperref}

\usepackage[numbers,sort]{natbib}
\usepackage{subfigure}
\usepackage{amsfonts}
\usepackage{booktabs}
\usepackage{tabularx}
\usepackage{cleveref}

\usepackage[disable, textsize=tiny]{todonotes}

\usepackage[inline]{enumitem}
\usepackage{wrapfig}

\newcommand{\R}{\mathbb{R}}
\newcommand{\I}{\mathbb{I}}
\newcommand{\ul}[1]{\underline{#1}}
\newcommand{\lb}[1]{\underline{#1}}
\newcommand{\ol}[1]{\overline{#1}}
\newcommand{\ub}[1]{\overline{#1}}
\newcommand{\iv}[1]{\underline{\overline{#1}}}

\newcommand{\method}{FullCert}

\newcommand{\twomoons}{Two-Moons}
\newcommand{\mnist}{MNIST 1/7}

\newif\ifreview
\reviewfalse

\ifreview
	\usepackage{lineno}

	\linenumbers
\fi

\begin{document}

\def\SubNumber{39}

\def\GCPRTrack{Fast Review Track}

\title{\method: Deterministic End-to-End Certification for Training and Inference of Neural Networks}

\ifreview
\else
	\titlerunning{\method: Deterministic End-to-End Certification of Neural Networks}

	\author{Tobias Lorenz\inst{1,2} \and
	Marta Kwiatkowska\inst{2} \and
    Mario Fritz\inst{1}
    }

	\authorrunning{T. Lorenz, M. Kwiatkowska, M. Fritz}
	
	\institute{CISPA Helmholtz Center for Information Security, Saarbr\"ucken, Germany \and Department of Computer Science, University of Oxford, Oxford, UK \\
    \smallskip
    \email{tobias.lorenz@cispa.de}, \email{marta.kwiatkowska@cs.ox.ac.uk}, \email{fritz@cispa.de}}
\fi

\maketitle              %

\begin{abstract}
Modern machine learning models are sensitive to the manipulation of both the training data (poisoning attacks) and inference data (adversarial examples).
Recognizing this issue, the community has developed many empirical defenses against both attacks and, more recently, certification methods with provable guarantees against inference-time attacks.
However, such guarantees are still largely lacking for training-time attacks.
In this work, we present \method, the first end-to-end certifier with sound, deterministic bounds, which proves robustness against both training-time and inference-time attacks.
We first bound all possible perturbations an adversary can make to the training data under the considered threat model.
Using these constraints, we bound the perturbations' influence on the model's parameters.
Finally, we bound the impact of these parameter changes on the model's prediction, resulting in joint robustness guarantees against poisoning \emph{and} adversarial examples.
To facilitate this novel certification paradigm, we combine our theoretical work with a new open-source library BoundFlow, which enables model training on bounded datasets.
We experimentally demonstrate \method's feasibility on two datasets.
\end{abstract}

\section{Introduction}
\label{sec:introduction}

Deep neural networks have shown impressive performance across many tasks, especially computer vision \citep{dosovitskiy2021visontransformers} and natural language \citep{brown2020language} tasks.
However, current research into the robustness of these models also reveals that their heavy reliance on data makes deep learning models susceptible to both training-time and inference-time data attacks \citep{papernot2016science}.
Training-time attacks, e.g., data-poisoning \citep{schwarzschild2021just, zhong2020backdoor}, manipulate the \emph{training} data to change the model and, therefore, its predictions.
Inference-time evasion attacks, e.g., adversarial examples \citep{goodfellow2015explaining, szegedy2014intriguing}, change the \emph{test} data to alter the prediction.
To achieve reliable, trustworthy machine learning models, it is crucial to quantify and limit an attacker's influence on the models and to develop worst-case guarantees on the model's performance.
Robustness guarantees will only become more critical as we increase the model's scale and train them on massive datasets scraped from the internet \cite{brown2020language, chatgpt}.

\begin{figure}
    \centering
    \includegraphics[width=0.7\columnwidth]{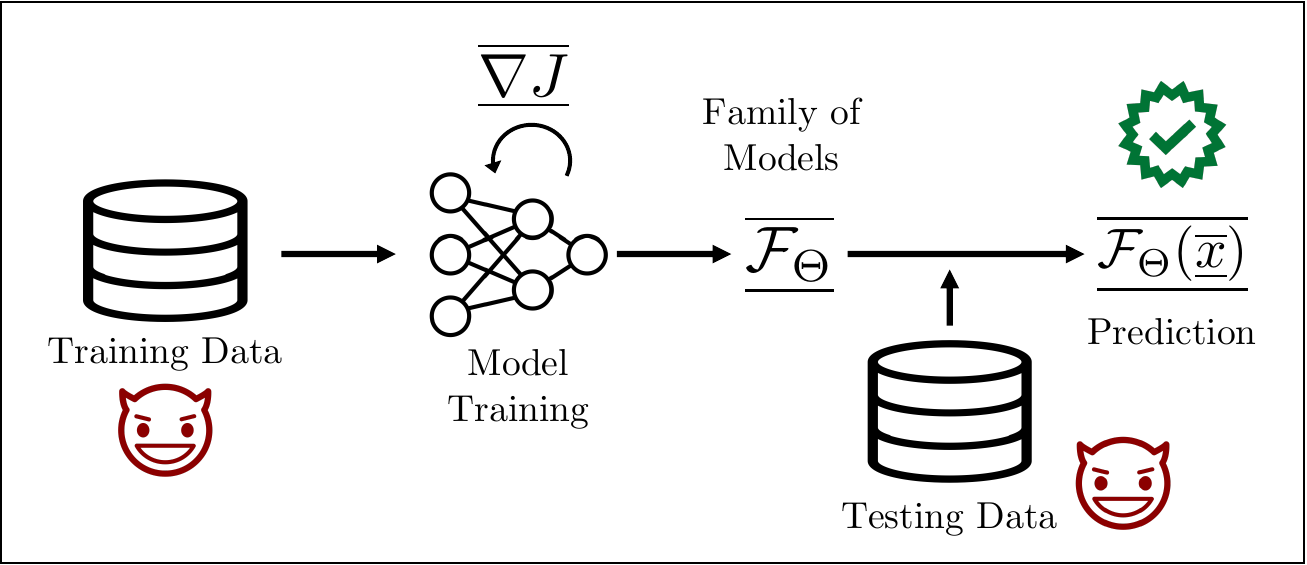}
    \caption{Overview of \method. During training, we bound the effects that perturbation of the training data can have on the model. At inference, we combine these bounds with bounds for perturbation of the test data. The result is a certified prediction against training {\it and} test time attacks.}
    \vspace{-.5em}
    \label{fig:teaser}
\end{figure}

Certified defenses against evasion attacks, i.e., attacks to change the behavior of a pre-trained model, have made significant progress over the last years, yielding solutions that can guarantee the robustness of medium-sized models \cite{carlini2022certified, ferraricomplete}.
In contrast, for training-time poisoning defenses, current solutions still largely rely on heuristics without any rigorous guarantees and can therefore be circumvented by adaptive attacks \citep{koh2022stronger}.
First works on certified defenses against poisoning attacks with probabilistic guarantees \citep{jia2022certified, rosenfeld2020certified, wang2020certifying, zhang2022bagflip} show that the problem is much harder, requiring further investigation and new solutions.

Multiple dimensions make training-time certification more difficult than in\-fer\-ence-time certification.
While inference only depends on a single data point, training depends on a large set of training data.
This gives the attacker a much higher degree of freedom, as they can influence multiple data points concurrently.
A second aspect is the much deeper computational graph.
While inference consists of a single forward pass, training consists of multiple iterations of forward and backward passes for each data point and, usually, for multiple epochs.
Since certifying the robustness of a single forward pass is already NP-complete \citep{katz2017reluplex}, certifying the robustness of the entire training procedure becomes even more challenging.
Current approaches give only probabilistic guarantees and severely limit the attacker's influence to a few data samples \cite{jia2021intrinsic, rosenfeld2020certified} or single pixels \citep{wang2020certifying}.

We propose \method{} -- the first end-to-end certifier with deterministic worst-case guarantees against both training-time poisoning and inference-time evasion attacks.
\method{} consists of three elements: (1) a formal problem definition of deterministic end-to-end certification, (2) a deterministic, sound certification formulation, and (3) an instantiation and implementation using interval bounds based on our new BoundFlow library.
Our approach is based on abstract interpretation, which uses reachability analysis to train a family of infinitely many models that could have resulted from all possible poisonings within some bound of the training data.
We then perform inference on this family of models, considering all possible input perturbations.
This allows us to determine whether the {\it combination} of poisoning and evasion attack could have affected the final prediction and establish a robustness certificate if we can guarantee correctness.
We implement our certifier in an open-source software package BoundFlow based on PyTorch and experimentally validate it on different tasks. We summarize our contributions as follows:
\begin{enumerate*}
    \item A formal definition of the deterministic end-to-end neural network certification problem
    \item \method, the first deterministic certifier against both data-poisoning and evasion attacks
    \item A formal convergence analysis of our method
    \item An open-source software package BoundFlow
    \item An experimental evaluation of \method.
\end{enumerate*}
Our implementation is available at \url{https://github.com/t-lorenz/FullCert}.

\section{Related Work}
\label{sec:related-work}
The work most closely related to our method stems from two general lines: first approaches to probabilistic robustness certification of training-time data-poisoning attacks and sound, deterministic certifiers against evasion attacks.

\paragraph{Training-Time Certification.}
\label{sec:related-work:training}

The most closely related work to our approach is \citet{wang2020certifying}, which proposes a defense against backdoor attacks via Randomized Smoothing.
By smoothing over training subsets, they are able to extend randomized smoothing to network training.
Using a subset of 100 digits from two MNIST classes, converted into black-and-white values, the method can guarantee invariance to the modification of 2 pixels across all training images and the test image.
In contrast, our guarantees are deterministic and hold with certainty, and we consider imperceptible perturbations to \emph{all} pixels.

\citet{rosenfeld2020certified} also use Randomized Smoothing for guarantees against label-flipping attacks, and \citet{weber2023rab} extend Randomized Smoothing to defend against backdoor attacks with probabilistic guarantees to $\ell_2$-norm perturbations.

A second line of work derives (probabilistic) guarantees against training-time attacks through bagging.
\citet{jia2021intrinsic} demonstrate that the data sub-sampling of vanilla bagging strategies show intrinsic robustness against poisoning attacks.
\citet{jia2022certified} show that this property natively holds for nearest neighbor classifiers, even without ensembles.
\citet{wang2022improved} build upon the bagging approach and improve the robustness guarantees based on an improved sampling strategy.
\citet{levine2021deep} propose a deterministic version of bagging, which partitions the dataset based on a deterministic hash function.
\citet{zhang2022bagflip} extend this approach to consider backdoor attacks with triggers.
Recent work also considers different threat models, including temporal aspects \cite{wang2023temporal} and dynamic attacks \cite{bose2024certifying}.

The main differences between this line of work and our approach are the (mostly) probabilistic versus our deterministic guarantees and the limitation to training-time attacks, whereas ours enables end-to-end certification with in\-fer\-ence-time attacks, and different threat model assumptions ($\ell_0$/$\ell_2$ vs. $\ell_\infty$).
We include a detailed comparison between the two approaches in \cref{sec:experiments}.

\paragraph{Inference-Time Certification.}
\label{sec:related-woark:inference-time}
There is a long list of work on bound-based test- and inference-time certification \citep{boopathy2019cnn, gehr2018ai2, mirman2018differentiable, singh2019abstract, weng2018towards, zhang2018efficient}.
The main difference is their trade-off between the precision of the certificate versus the scalability of the method to larger network sizes.
Most closely related to our work is Interval Bound Propagation (IBP) \citep{gowal2018effectiveness}, which, like our approach, uses intervals with upper and lower real-valued bounds.

\paragraph{Bound-Based Robust Training.}
\label{sec:related-woark:robust-training}
Many inference-time certification approaches also exploit their bounds during training to achieve more robust models \citep{gowal2018effectiveness, mirman2018differentiable, zhang2018efficient}, sometimes referred to as ``robust training'' or ``certified training'' in the literature, which may cause confusion with our method.
The goal of using bounds for training is to improve the model's robustness against \emph{inference-time} attacks and to mitigate over-approximations.
In contrast to our work, they do not provide any defense or guarantees against \emph{training-time} attacks.

\section{End-to-End Neural Network Certification}
\label{sec:certification}

We present our method \method, the first deterministic end-to-end neural network certifier against training-time and inference-time attacks.
In \cref{sec:certification:problem}, we formally define the end-to-end certification problem.
We then present our general approach based on reachability analysis (via abstract interpretation) in \cref{sec:certification:general}, independent of the concrete bounds.
Finally, we introduce our instantiation using interval bounds in \cref{sec:certification:intervals} and conclude with a description of our implementation in \cref{sec:certification:implementation}.

\subsection{Problem Definition of End-to-End Certification}
\label{sec:certification:problem}

The goal of \method{} is to certify the robustness of a deep learning model against perturbations on the training and inference data.
Intuitively, we view the combination of the model training and then its prediction on a single data point as a single function.
This function takes a training dataset $D_\text{train}$ consisting of $N$ pairs of inputs and labels, as well as a single test sample $x$ as input.
We can then analyze the influence of changing a combination of $D_\text{train}$ and $x$ on the model's prediction.
If we can guarantee that no changes would alter the prediction, we can certify the model's robustness.

\paragraph{Certified Training.}
More formally, we train a model $f_\theta: \mathcal{X} \mapsto \mathcal{Y}$ on a dataset $D$ by optimizing its parameters $\theta$ using a training algorithm $A$, e.g., SGD:
\begin{equation}
    \theta = A(D).
\end{equation}
An adversary can influence the training set with a poisoning attack, which we model by considering a family of datasets $\mathcal{D}$, which include all bounded perturbations that the adversary could cause to the training data.
One way to define $\mathcal{D}$ is through a similarity metric $d$, which measures the similarity of two datasets:
\begin{equation}
    \mathcal{D} := \left\{D \mid d(D, D_\text{train}) \leq \epsilon \right\}.
\end{equation}
We use the element-wise $\ell_\infty$-norm with small $\epsilon$ to restrict the adversary to imperceptible changes in analogy to evasion attacks.

Given $\mathcal{D}$ as possible inputs to $A$, we get a family of models
\begin{equation}
    \mathcal{F}_\Theta := \left\{ f_\theta \mid \theta = A(D), D \in \mathcal{D} \right\}
\end{equation}
which contains all possible models that could result from the poisoning attacks.
Assuming no inference-time attacks after deployment, we can certify the model robustness at $x$ if
\begin{equation}
    \forall f_\theta \in \mathcal{F}_\Theta : f_\theta(x) = y.
    \label{eq:training-certificate}
\end{equation}

\paragraph{End-to-End Certification.}
For end-to-end certification, we also need to consider perturbations to $x$.
We adopt the common formulation of adversarial examples \cite{goodfellow2015explaining}, and define the bounded perturbation set $\mathcal{S} := \{x' \mid d'(x, x') \leq \epsilon'\}$.
$d'$ measures the distance between the original and perturbed data point, for which we use the $\ell_\infty$-norm again.
We can then extend \cref{eq:training-certificate} to certify end-to-end robustness against both training and inference time attacks:
\begin{equation}
    \forall f_\theta \in \mathcal{F}_\Theta, x' \in \mathcal{S} : f_\theta(x') = y.
    \label{eq:full-certification}
\end{equation}

\subsection{\method{}: End-to-End Certification of Neural Networks by Abstract Interpretation}
\label{sec:certification:general}

Solving \cref{eq:full-certification} precisely is infeasible, as even the test-time certification problem for a single classifier, that is, $\forall x' \in \mathcal{S} : f_\theta(x') = y$, is NP-complete \citep{katz2017reluplex}.
We, therefore, propose a solution inspired by work on test-time certification, which uses reachability analysis (based on abstract interpretation \citep{cousot1977abstract}) to compute a solution.
The key is to only allow over-approximations in order to ensure sound certificates.
In other words, if we guarantee robustness (i.e., the certifier outputs that \cref{eq:full-certification} holds), we want to be sure this guarantee is correct.
However, we allow the certifier to not compute a guarantee in some cases, even though the underlying model is actually robust, to make a solution feasible.

To perform this reachability analysis, we view the training of the model on a dataset $D$ and the following inference on a single datapoint $x$ as a single function, which takes $D$ and $x$ as inputs and computes the prediction $f(x)$ as output.
This allows us to define the potential perturbations on $D$ and $x$ as a precondition, propagate it through the unrolled training and inference, and finally check the postcondition of correct classification.

The first step in this process is to build the sets $\mathcal{D}$ and $\mathcal{S}$, which contain all possible perturbations permitted by the attack model.
The concrete shape of these sets depends on the threat model, which in our case is an interval (also called hyperrectangle, orthotope, or box) with real-valued upper and lower bounds.

Once $\mathcal{D}$ and $\mathcal{S}$ are defined, we compute all outputs that could result from any combination of elements from those input sets.
Since both sets contain infinitely many elements, we cannot test all combinations.
We therefore unroll the training and subsequent inference into a series of function invocations
$g_1 \circ g_2 \circ \cdots \circ g_n$, where each $g_i$ represents a part of a forward or backward pass.
For example, the layers of the forward pass, the loss function, or the gradient computations of the backward pass.
We can then define abstract versions of $g_i$, labeled $G_i$, which compute the effect of $g_i$ on an entire set of inputs.

More formally, given a function $g_i: \mathcal{U} \to \mathcal{V}$, we define an abstract version for sets $G_i: \mathbb{U} \to \mathbb{V}$, where $U \in \mathbb{U}, U \subset \mathcal{U}$, such that
\begin{equation}
    \forall u \in U, U \in \mathbb{U} : g_i(u) \in G_i(U).
    \label{eq:over-approximation}
\end{equation}
This property ensures that $G_i(U)$ contains all reachable points of $g_i(u)$, which makes the method sound\todo{soundness as ``definition''}.
The opposite does not necessarily hold, i.e., we allow points in $G_i(U)$ that are not reachable by $g_i(u)$\todo{soundness of chain of functions implies soundness of the entire function as corollary}.
This over-approximation makes the computation feasible, as it is not always possible to efficiently compute an exact $G_i(U)$.
The next paragraphs describe training and inference in detail.

\paragraph{Forward Pass.}
For each training sample and its ground-truth label $(x, y) \in D_\text{train}$, the first step is to build the set $\mathcal{X}_0 = \{x' \mid d(x', x) \leq \epsilon\}$.
We then execute the forward pass on $\mathcal{X}_0$ by building and abstract operation $L^{(i)}$ for each layer operation $l^{(i)}$ in $f_\theta = l^{(1)} \circ \cdots \circ l^{(k)}$, as detailed in \cref{eq:over-approximation}.
After the last layer, we get a set $\mathcal{X}_k = L^{(k)}(\mathcal{X}_{k-1})$ with possible outputs.
We compute the set of losses $\mathcal{O}$ on $\mathcal{X}_k$ using an abstract version of the loss function via the same principle.

\paragraph{Backward Pass.}
The backward pass uses the chain rule to compute the gradients wrt each parameter.
Since all layer inputs and outputs are sets, we have to create abstract versions of the gradient functions for each layer.
This means the gradient with respect to each parameter is a set of values as well.
This set represents the influence an attacker has on the gradient by perturbing the training data.
With sets as gradients, the parameter update also turns into a set operation:
\begin{equation}
    \Theta_{t+1} = \Theta_t - \lambda_t \nabla_\theta J(\Theta_t, X, y),
    \label{eq:gradient-update}
\end{equation}
with objective function $J$ and learning rate $\lambda_t$.
This leads to a set of possible parameters $\Theta$, representing all models that could result from training with perturbations.
This yields a multiplication of two sets, the parameters and the inputs, in all linear and convolution layers, which is an additional challenge compared to inference-time-only certification without bounds on the parameters.

The final result after this training is a family of models $\mathcal{F}_\Theta$, represented by a model architecture parameterized with an infinite set of parameters. This set of parameters contains all possible parameters the attacker could impose through data poisoning, in addition to elements due to over-approximation.

\paragraph{Inference.}
Using these sets of parameters resulting from training, we compute the final certificate during inference for a test input $x$.
The principle is the same as a forward pass during training: build the set $\mathcal{X}_0$, and then propagate it through the abstract operations to obtain a set of outputs $\mathcal{X}_k$.
If all of these possible outputs are classified correctly, we can conclude that no attack could have changed the prediction.

This certification scheme can be used both as online certificates, guaranteeing a robust prediction for a given sample, and as offline certificates, quantifying the robustness of a model over a test set.

\subsection{\method{} Based on Interval Abstractions}
\label{sec:certification:intervals}

The general solution introduced in \cref{sec:certification:general} is independent of the concrete choice of how to represent the sets and abstract functions.
The literature for inference-time certification has used different convex models, from fast intervals to more complex polytopes.
The choice of relaxation controls the trade-off between computational complexity and the precision of the over-approximations.

For this work, we use intervals (or hyperrectangles/boxes) for our relaxations.
This abstraction has two main advantages: (1) interval bounds are fast to compute compared to other relaxations, which is crucial due to the large depth of the computational graph for model training, and
(2) it is easy to implement the multiplication of two intervals, which is not straightforward for more general polytopes.
We introduce abstract interval-versions of the operations required for neural network training below, and prove their soundness in \cref{app:relaxations}.

\paragraph{Notation.}
We represent each set of values, e.g., $\mathcal{X}$ and $\Theta$, as an $n$-dimensional interval with upper and lower real bounds.
We denote an interval as $\iv{A} = [\ul{A}, \ol{A}] \in \I^n$, with lower bound $\ul{A} \in \R^m$ and upper bound $\ol{A} \in \R^m$.
Equivalently, we can define the same interval via its center and radius: $\iv{A} = \{m_A, r_A\}$, where $m_A = \frac{\ul{A} + \ol{A}}{2}$ is the center and $r_A = \frac{\ol{A} - \ul{A}}{2}$ the radius.
We use these interchangeably.

\paragraph{Linear Layer.}
For linear and convolution layers, we use Rump's algorithm~\cite{diep2010efficient, rump1999fast} for matrix multiplication based on the center and radius of the interval:
\begin{equation}
    \begin{split}
    &\iv{C} = \iv{A} * \iv{B}, \iv{A} \in \I^{m \times n}, \iv{B} \in \I^{n \times p}, \iv{C} \in \I^{m \times p}\\
    &m_C = m_A * m_B,\ r_C = (|m_A| + r_A) * r_B + r_A * |m_B|.
    \end{split}
\end{equation}

We can compute bounds for linear layers with weights $\iv{W}$, biases $\iv{B}$, and inputs $\iv{X}$, as well as the derivatives:
\begin{equation}
    \iv{Y} = \iv{W} \iv{X} + \iv{B}, \qquad
    \iv{\frac{\partial Y}{\partial X}} = \iv{W} \quad
    \iv{\frac{\partial {Y}}{\partial {W}}} = \iv{X} \quad
    \iv{\frac{\partial {Y}}{\partial {B}}} = \mathbf{1}.
\end{equation}

\paragraph{Activation Functions.}
We can compute the upper and lower bound for any monotonic function by evaluating only the interval bounds (proof in \cref{app:relaxations:monotonic}):
\begin{equation}
    G(\iv{X}) = [\min\{g(\ul{X}), g(\ol{X})\}, \max\{g(\ul{X}), 
    g(\ol{X})\}].
\end{equation}
In particular, the ReLU function and its derivative can be bound as:
\begin{equation*}
    \text{ReLU}(\iv{X}) = [\text{ReLU}(\ul{X}), \text{ReLU}(\ol{X})], \qquad \text{ReLU}'(\iv{X}) = [\text{ReLU}'(\ul{X}), \text{ReLU}'(\ol{X})].
\end{equation*}

\paragraph{Cross-Entropy Loss.}
To compute the gradients for the backward pass, we first need an interval version of the cross-entropy (CE) loss function.
CE is defined as $J(p, y) = -\sum_{i=1}^m{y_i \log(p_i)}$ on the class probabilities $p_i$.
A naive solution could use the upper and lower bounds of the softmax function followed by the logarithm.
However, this is numerically unstable and, therefore, the softmax and logarithm are typically computed  together~\cite{goodfellow2016deep}.
We take this into account and define numerically stable bounds for the logsoftmax function:
\begin{equation}
\begin{split}
    \mathrm{logsoftmax}(\iv{z}) =
    \Biggl[
        (\ul{z}_c - a) - \log\Biggl(\sum_{i \neq c}^m{\exp(\ol{z}_i - a)} + \exp(\ul{z}_c - a) \Biggr),\\
         (\ol{z}_c - b) - \log\Biggl(\sum_{i \neq c}^m{\exp(\ul{z}_i - b)} + \exp(\ol{z}_c - b) \Biggr) \Biggr]\\
     a = \max\{\ul{z}_c, \ol{z}_{i \neq c}\},\  b = \max\{\ol{z}_c, \ul{z}_{i \neq c}\}.
\end{split}
\end{equation}
Refer to \cref{app:relaxations:crossentropy} for a detailed derivation, proofs of the soundness and tightness of these bounds, and numerically stable bounds for the CE derivative.

\paragraph{Binary Cross-Entropy.}
The BCE loss with sigmoid function $\sigma(z)$ is
\begin{equation}
    J(p, y) = -(y \log(p) + (1 - y) \log (1 - p)),\ p = \sigma(z).
\end{equation}
It has similar stability issues as CE with softmax.
We define its bounds as
\begin{equation}
\begin{split}
    J(\iv{z}, y) = \bigl[\lb{z} - \lb{z}y + a + \log(\exp(-a) + \exp(-\lb{z} -a)),\\
    \ub{z} - \ub{z}y + b + \log(\exp(-b) + \exp(-\ub{z}-b))\bigr]\\
    a=\max(-\lb{z}, 0),\ b=\max(-\ub{z}, 0)
\end{split}
\end{equation}
and defer the derivative and proofs to \cref{app:relaxations:bce}.

\subsection{Implementation}
\label{sec:certification:implementation}

Existing bound-based certifiers only consider inference-time certification.
They rely on standard deep learning libraries for training, which use tensors as their basic data type and cannot be easily extended to support bounded parameters.
We, therefore, develop a new open-source software package for end-to-end certification.
The primary goals are (1) a correct implementation of our method, (2) easy extensibility, (3) high performance through GPU support, and (4) easy integration with existing deep-learning frameworks.

We choose PyTorch's \cite{Paszke2019pytorch} low-level vector operations as our basis, which allows for performance-optimized operations.
However, we cannot re-use PyTorch's pre-defined layer operations or automatic differentiation system, as it assumes tensor inputs.
We, therefore, create our own open-source library BoundFlow, which implements layers, models, and gradient operations for bound-based model training, including GPU support, for many platforms.
It supports basic layer operations for interval bounds, which can be extended to more complex operations and bounds as required.

\section{Convergence Analysis}
\label{sec:convergence}

Training with \method{} can be thought of as executing SGD on infinitely many datasets concurrently.
We analyze the convergence of \method{} towards an optimum to understand under which conditions certified training can work and where the challenges lie.
Since proving the convergence of SGD for general, non-convex problems is not possible, we consider a simplified, convex setting and full-batch gradient descent, which is well understood \citep{bottou1998online} and gives insights into the algorithm's behavior.

In principle, two factors can cause the training to diverge. (1) The perturbations of the training set itself, i.e., the perturbations could include a dataset for which gradient descent no longer converges. (2) The over-approximations, which we allow to make the solution feasible, could prevent the algorithm from converging in some cases.
We consider both factors in turn.

\paragraph{Perturbations.}
To analyze the influence of perturbations, we consider a hypothetical, exact certifier, which does not allow any over-approximations.
That is, $\forall v \in G_i(U)\ \exists u \in U : g_i(u) = v$.
Then we can show that the certified training $A_\text{exact}$ on the family of perturbed datasets $\mathcal{D}$ converges exactly when the base gradient descent (GD) algorithm $A_\text{GD}$ converges for each perturbed dataset:
\begin{equation}
    \begin{split}
        \Theta_\text{exact} = A_\text{exact}(\mathcal{D}) \quad
        \forall \theta \in \Theta_\text{exact} : \exists D \in \mathcal{D} \mid \theta = A_\text{GD}(D)\\
        A_\text{GD} \text{ converges } \forall D \in \mathcal{D} \Leftrightarrow A_\text{exact}(\mathcal{D}) \text{ converges}.
    \end{split}
\end{equation}
This convergence property directly follows from the definition of exact certifiers, which implies that all parameter configurations $\theta \in \Theta_\text{exact}$ are reachable.
Therefore, there has to be a perturbed dataset $D \in \mathcal{D}$ that produces the same $\theta$ using $A_\text{GD}$.
The convergence property is desirable, as it implies a guarantee that the attacker could not have changed the convergence if certified training converges.

\paragraph{Over-Approximations.}
Practical certifiers cannot compute $\Theta_\text{exact}$ and therefore typically approximate it with $\Theta_\text{relaxed} \supset \Theta_\text{exact}$, which allows new values that could lead to divergence.
We analyze this with a proof sketch following \citet{bottou1998online}, which we generalize to families of datasets and sets of parameters.

As mentioned above, this proof only holds with some assumptions: (1) non-zero gradients everywhere except for the optimum (i.e., a single global optimum without saddle points), (2) Lipschitzness (i.e., bounded gradients), and (3) a decreasing, non-zero learning rate.

The core idea of \citet{bottou1998online} is to define a Lyapunov sequence $h_t = (\theta_t - \theta^*)^2$, a sequence of positive numbers whose value measures the distance to the target at step $t$.
We can then show that the algorithm converges by showing that this sequence converges.
We generalize this sequence to intervals as $h_t = (\Theta_t - \Theta^*)^2$.

Using the definition of the gradient update step (\cref{eq:gradient-update}), we get
\begin{equation}
    h_{t+1} - h_t = \underbrace{-2 \lambda_t (\Theta_t\color{black} - \Theta^*)\nabla_\Theta J(\Theta_t)}_\text{distance to optimum} + \underbrace{\lambda_t^2 (\nabla_\Theta J(\Theta_t))^2}_\text{discrete dynamics}.
\end{equation}
The second term is a consequence of the discrete nature of the system and is bounded by assumptions on the gradients and learning rate (see \citet{bottou1998online}).
The first term relates to the distance to the optimum and should be negative for the training to converge.
While this is easy to show for single $\theta_t$ and gradients, it does not always hold for intervals.
For the term to be negative, $(\Theta_t\color{black} - \Theta^*)\nabla_\Theta J(\Theta_t)$ has to be positive.
This is the case exactly when $\Theta_t \cap \Theta^* = \emptyset$ due to the properties of interval multiplication.
The algorithm starts diverging as soon as the optimum is contained within the parameter interval.

The second challenge lies in the gradient update step, which causes the radius of $\Theta_t$ to always grow.
This is a consequence of subtraction in interval arithmetic:
\begin{align}
    \iv{c} = \iv{a} - \iv{b} = [\underline{a} - \overline{b}, \overline{a} - \underline{b}] \Rightarrow r_c = r_a + r_b, \quad \Rightarrow
    r_{\Theta_{t+1}} = r_{\Theta_{t}} + \lambda r_{\nabla_\theta L(\Theta_t)}.
\end{align}

We draw two conclusions from this. (1) The fewer steps are required, the more precise the solution will be.
This is partially due to fewer over-approximations but also a natural consequence of the fact that, the less we train, the smaller the influence of an attacker.
(2) A small step size is beneficial, as the parameters slowly converge towards the optimum without including it within the parameter set prematurely.

\section{Experiments}
\label{sec:experiments}

To complement our theoretical insights, we perform a series of experiments to demonstrate the feasibility of \method.
We evaluate \method{} qualitatively and quantitatively on two different datasets.
\Cref{app:experiments} contains additional experiments, e.g., on the influence of hyper-parameters and model architectures.

\paragraph{Experimental Setup.}
\begin{wrapfigure}{r}{.5\linewidth}
    \vspace{-1em}
    \centering
    \includegraphics[width=\linewidth]{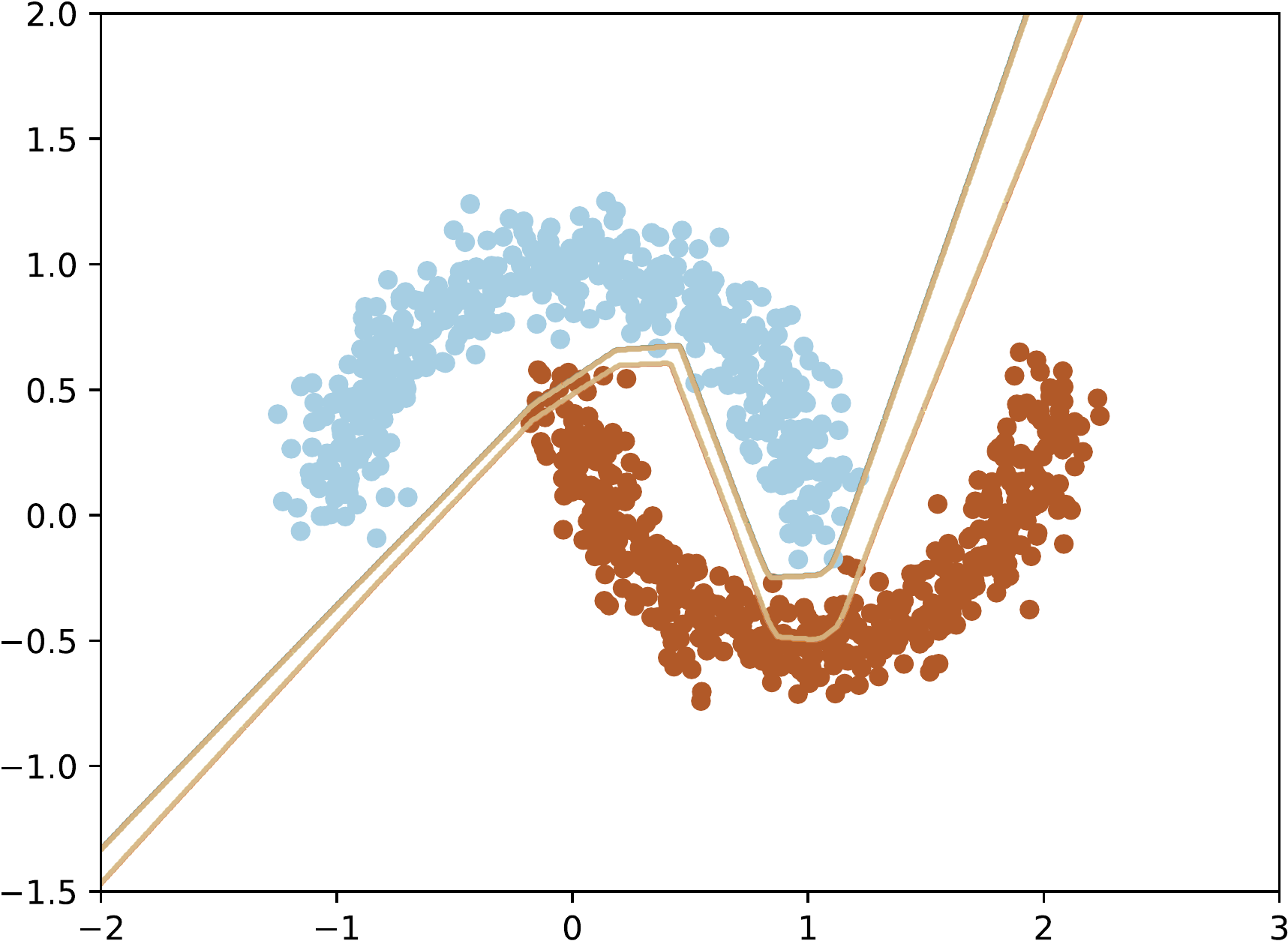}
    \caption{
        Barriers to the decision boundaries for all models that could have resulted from poisoning the \twomoons\ dataset.
        Our bounds on the parameters guarantee that all points outside these barriers are robustly classified.
    }
    \vspace{-2em}
    \label{fig:decision-bounds}
\end{wrapfigure}
We use our Bound\-Flow software package to implement end-to-end certified training and inference.
For evaluation, we use two datasets: \twomoons{} and \mnist{}~\cite{lecun1998gradient}.
\twomoons{} is a two-dimensional dataset with two classes of points configured in interleaving half circles.
It is well-suited for analyzing the behavior of training algorithms and allows easy visualization.
To test the behavior on more complex data, we follow prior work \citep{wang2020certifying} and train a model on \mnist, the MNIST subset with only digits 1 and 7.
We report \emph{certified accuracy}, the percentage of test samples for which we can guarantee correct prediction.
That means the model predicts the correct label, and \cref{eq:full-certification} holds.
We set the perturbation radii to $\epsilon = \epsilon' = 10^{-3}$ for \twomoons{} and $\epsilon = \epsilon' = 10^{-4}$ for MNIST.
Training details are listed in app.~\ref{app:experiments}, and our implementation is available at \url{https://github.com/t-lorenz/FullCert}.

\paragraph{End-to-End Certification.}
\label{sec:experiments:main}
These experiments demonstrate \method{} for SGD training.
The plethora of successful poisoning attacks shows that SGD training is unstable and easy to influence, which makes certification challenging.

To illustrate \method, we perform a qualitative evaluation on \twomoons{}.
\Cref{fig:decision-bounds} visualizes the dataset's two classes and the barriers to the decision boundaries of the model family trained with \method{}.
\method{} guarantees that the decision boundaries of all models lie between the two barriers, and therefore guarantees that all points outside these barriers are robustly classified.
An adversary could potentially have influenced the classification of points in between.

For a quantitative evaluation, we run experiments on \twomoons{} and \mnist{} datasets.
We start certified training from different starting points to verify the results from \cref{sec:convergence}.
Using a small subset of 100 unperturbed samples, we gradually pretrain models to different starting accuracies.

\begin{figure}
    \centering
    \vspace{-1em}
    \subfigure[\twomoons]{\label{fig:mdleft}{\includegraphics[width=.495\columnwidth]{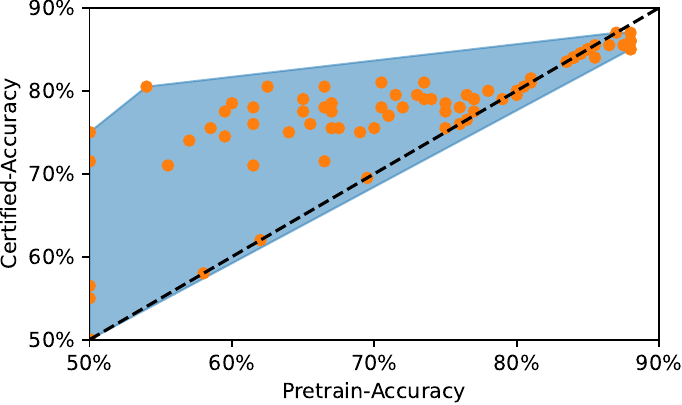}}}
    \subfigure[\mnist]{\label{fig:mdright}{\includegraphics[width=.495\columnwidth]{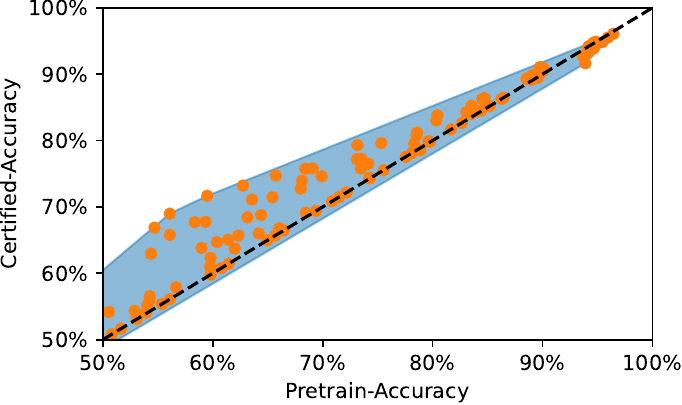}}}
    \vspace{-1em}
    \caption{
        Certified accuracy for different initial accuracies on \twomoons\ for $\epsilon = 10^{-3}$ and \mnist\ for $\epsilon = 10^{-4}$.
        Each dot represents a separate model, with the convex hull in blue.
        The closer the initialization after pretraining is to the final operating point, the higher the final certified accuracy.
        Blue convex hulls are for visualization purposes.
    }
    \vspace{-1.5em}
    \label{fig:pretrain_certification}
\end{figure}

\Cref{fig:pretrain_certification} shows the certified accuracy for both datasets for different starting points.
The diagonal in the center represents the baseline certified accuracy with only pretraining.
Points above the diagonal show improved certified accuracy, demonstrating benefits of our training scheme.

The results show that certified training is already effective without any pretraining (left end of the x-axis with $50\%$ accuracy).
On \twomoons, it achieves $75\%$ certified accuracy when training from scratch, and $60\%$ on \mnist.
The closer the model gets to the optimum during pretraining, the higher the final certified accuracy, slowly converging with the pretrained accuracy once the starting point reaches the final optimum.

\paragraph{Comparison to Related Approaches.}
\label{sec:experiments:baseline}
In contrast to certification for inference, training-time certification is still in its early stages.
In fact, our submission addresses deterministic end-to-end certification for the first time and is the first work to bring bound-based certification to training-time guarantees.

\begin{figure}
    \centering
    \vspace{-1em}
    \includegraphics[width=\columnwidth]{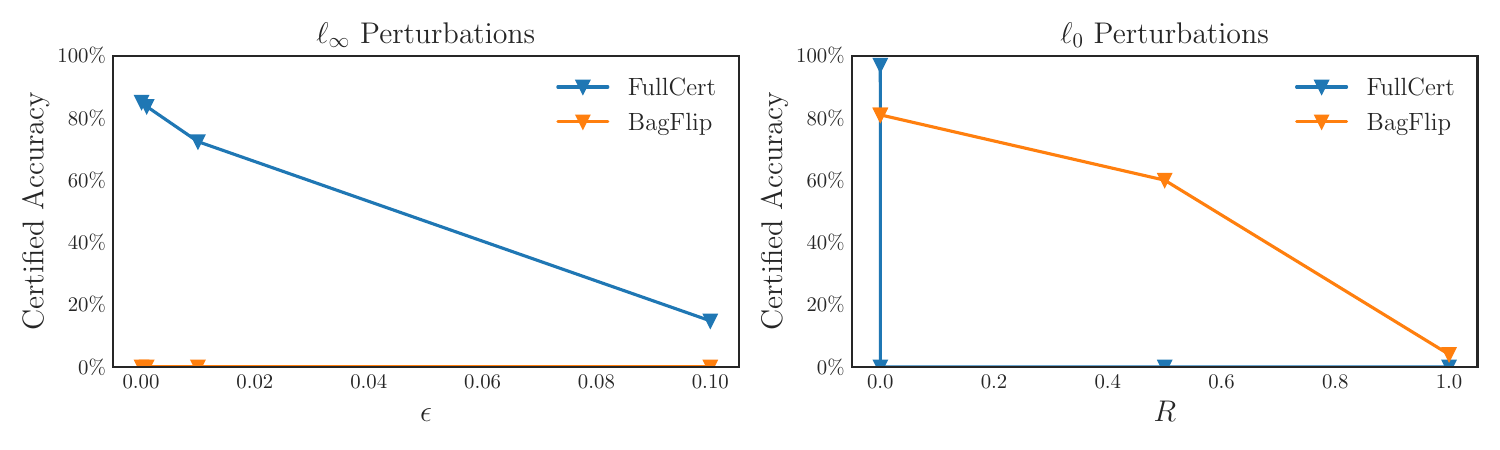}
    \vspace{-2.5em}
    \caption{Comparison between \method{} and BagFlip.
    Left: Certified Accuracy for different $\epsilon$ on Two-Moons. The threat model allows perturbations of up to $\epsilon$ for each feature.
    Right: Certified Accuracy for different $R$ as a percentage of MNIST 1/7 images with one flipped feature or label (FL1 perturbation) as reported in BagFlip.
    }
    \vspace{-1em}
    \label{fig:comparison}
\end{figure}

As discussed in \cref{sec:related-work:training}, existing work on training-time certification is based on bagging and randomized smoothing.
These are fundamentally different certification techniques with different assumptions on the threat model, which make a direct comparison to our approach impossible.
To illustrate this difference, we compare \method{} to BagFlip~\citep{zhang2022bagflip}. 
(1) BagFlip guarantees robustness against $\ell_0$-``norm'' perturbations, which means robustness against flipping a small percentage of features or labels of the training or test data. In contrast, FullCert guarantees robustness against $\ell_\infty$-norm perturbations, which means small changes to all features of the training and test data.
Ideally, machine learning models should be robust against both. 
\Cref{fig:comparison} illustrates that each method succeeds in its perturbation model, but does not transfer to the other.
(2) \method{} computes deterministic guarantees, which always hold.
In contrast, BagFlip’s guarantees only hold with high probability, which means a small chance remains that the guarantee is invalid.
(3) BagFlip requires training and evaluation of 1000 models for the same task. In contrast, \method{} only trains a single model with bound propagation.
(4) \method{} evaluates the robustness of a normal model as if it was trained regularly, while BagFlip changes the underlying model into an ensemble, which results in different predictions.
(5) BagFlip only works on discretized features, while \method{} can handle both discretized and standard, real-valued features. 
All of these differences make it impossible to compare the two methods directly.
We believe there is merit in exploring both techniques for training-time certification, analogous to how the community explores both approaches for test-time certification concurrently.

\section{Discussion \& Limitations}
\label{sec:limitations}

End-to-end certification of networks is a crucial yet difficult problem.
We propose \method, the first deterministic certifier, which also extends its guarantees to model training.
As any method addressing challenging problems, \method{} comes with some limitations.

One challenge common to all certification methods is limited scalability to larger models.
This is due to the high computational complexity of the certification problem.
Even a single inference pass is already NP-complete \citep{katz2017reluplex}, which is exacerbated for training-time certification with multiple forward and backward passes.
We address this using fast interval relaxations, which make the computation feasible at the cost of over-approximations.
These over-approximations are currently the main limiting factor of our approach, leading to small certified radii.
They could be addressed with more precise relaxations in future work.
However, even in its current form, \method{} can be used to assess the stability of training against perturbations in smaller-scale settings.

The theoretical guarantees and bounds of \method{} are sound, as shown in \cref{sec:certification}.
Due to the underlying PyTorch framework not supporting sound floating-point arithmetic,
the implementation may suffer from numerical imprecision.
We believe that the effects are negligible in practice,
which is supported by prior work with the same limitation \citep{boopathy2019cnn, weng2018towards, zhang2018efficient}.

Finally, the algorithm and threat model we consider are challenging.
The work on poisoning attacks shows that training is very sensitive to per\-tur\-ba\-tions \cite{tian2022comprehensive}.
The threat model gives the adversary control over all inputs, which is a worst-case assumption.
The training instability and the adversary's high degree of freedom limit current guarantees to small perturbation radii compared to test-time certificates.
We see great potential for tighter guarantees using more robust training algorithms and limiting the adversary's control to a subset of the data.

\section{Conclusion}
\label{sec:conclusion}
This work addresses the challenging problem of end-to-end robustness certificates against training-time and inference-time attacks.
We formally define the certification problem and propose \method, the first certifier that can jointly compute sound, deterministic bounds for both types of attacks.
The theoretical analysis shows the soundness and convergence behavior of this method.
Our new open-source library BoundFlow allows the implementation and experimental evaluation of \method, demonstrating its feasibility.

\section*{Acknowledgments}
This work was partially funded by ELSA - European Lighthouse on Secure and Safe AI funded by the European Union under grant agreement number 101070617, as well as the German Federal Ministry of Education and Research (BMBF) under the grant AIgenCY (16KIS2012), and Medizin\-informatik-Platt\-form ``Privatsph\"aren-sch\"utzende Analytik in der Medizin'' (PrivateAIM), grant number 01ZZ2316G.
MK received funding from the ERC under the European Union’s Horizon 2020 research and innovation programme (FUN2MODEL, grant agreement number 834115).

\bibliographystyle{splncs04nat}
\bibliography{main}

\clearpage
\onecolumn
\appendix
\section*{Appendix}
\section{Interval Relaxations}
\label{app:relaxations}

This section provides additional information on the constraints introduced in \cref{sec:certification:intervals}, as well as proofs for the soundness and tightness of these bounds.

\subsection{Basic Operations}
\label{app:relaxations:basic}

The most basic form of abstract operation on these sets is the standard interval arithmetic, with the same soundness concept as introduced before.
Given two intervals $\iv{A}, \iv{B} \in \I^m$, we can compute their sum and differences as
\begin{equation}
\begin{split}
    &\iv{A} + \iv{B} = [\ul{A} + \ul{B}, \ol{A} + \ol{B}]\\
    &\iv{A} - \iv{B} = [\ul{A} - \ol{B}, \ol{A} - \ul{B}].
    \label{eq:interval-addition}
\end{split}
\end{equation}

Similar interval operations can be defined for multiplication and division:
\begin{equation}
    \begin{split}
        &\iv{A} * \iv{B} = \bigl[\min\{\lb{A} * \lb{B}, \lb{A} * \ub{B}, \ub{A} * \lb{B}, \ub{A} * \ub{B}\},\\
        &\qquad \max\{\lb{A} * \lb{B}, \lb{A} * \ub{B}, \ub{A} * \lb{B}, \ub{A} * \ub{B}\}\bigr]\\
        &\frac{\iv{A}}{\iv{B}} = \iv{A} * \frac{1}{\iv{B}}, \qquad
        \frac{1}{\iv{B}} = \left[\frac{1}{\ub{B}} , \frac{1}{\lb{B}} \right], 0 \notin \iv{B}.
    \end{split}
\end{equation}

\subsection{Monotonic Functions}
\label{app:relaxations:monotonic}

Monotonically increasing or decreasing functions can be evaluated at the bounds of the input interval.
For a monotonically increasing function $f$:
\begin{equation}
    f(\iv{x}) = [f(\lb{x}), f(\ub{x})].
    \label{eq:monotonic-increasing}
\end{equation}
This directly follows from the monotonic property, i.e., $\forall a, b: a \leq b \Rightarrow f(a) \leq f(b)$:
\begin{align*}
    \forall u \in \iv{x}: u \leq \ub{x} &\Rightarrow f(u) \leq f(\ub{x})\\
    \lb{x} \leq u &\Rightarrow f(\lb{x}) \leq f(u)
\end{align*}
\qed

These bounds are also tight, as they are realized by the upper and lower bounds of the input.

For monotonically decreasing functions, the opposite holds with the same argument:
\begin{equation}
    f(\iv{x}) = [f(\ub{x}), f(\lb{x})].
    \label{eq:monotonic-decreasing}
\end{equation}

\subsection{Cross-Entropy Loss}
\label{app:relaxations:crossentropy}

The typical loss function to train classifiers is the cross-entropy-loss:
\begin{equation}
\begin{split}
    J(x, y) = -\sum_{i=1}^m{y_i \log(p_i)}\\
    p_i := \text{softmax}_i(z) = \frac{e^{z_i}}{\sum_{j=1}^m{e^{z_j}}}
\end{split}
\end{equation}

Computing the last-layer softmax function followed by the cross entropy loss is numerically unstable since $\exp(z) = \text{inf}\ \forall z > 88$ and $\exp(z) = 0.0\ \forall z < -104$.
This is already a challenge in regular model training, but bound-based training is especially sensitive.

Deep learning frameworks, therefore, combine the two operations into a single, numerically stable operation using the logsoftmax trick \citep{goodfellow2016deep}:
\begin{align}
    \begin{split}
        \text{softmax}_c(z) = \frac{\exp(z_c)}{\sum_{i=1}^m{\exp(z_i)}} = \frac{\exp(z_c - a)}{\sum_{i=1}^m{\exp(z_i - a)}}\\
    \end{split}\\
    \begin{split}
        \text{logsoftmax}_c(z) &= \log\left(\frac{\exp(z_c)}{\sum_{i=1}^m{\exp(z_i)}}\right)\\
        &= (z_c - a) - \log\left(\sum_{i=1}^m{\exp(z_i - a)}\right)\\
        a = \max_{i=1..m}{z_i}
    \end{split}
\end{align}

Na\"ive bounds for softmax would be
\begin{equation}
\begin{split}
    \text{softmax}_c(\iv{z}) = \left[ \frac{\exp(\ul{z}_c)}{\sum_{i=1}^m{\exp(\ol{z}_i)}}, \frac{\exp(\ol{z}_c)}{\sum_{i=1}^m{\exp(\ul{z}_i)}} \right]\\
    = \left[ \frac{\exp(\ul{z}_c - a)}{\sum_{i=1}^m{\exp({\ol{z}_i - a)}}}, \frac{\exp(\ol{z}_c - a)}{\sum_{i=1}^m{\exp(\ul{z}_i - a)}} \right]
\end{split}
\end{equation}
However, these bounds are not tight. The upper and lower bounds of $z_c$ are used in the same term but can never be realized simultaneously.
Also, using a single offset $a$ does not fully mitigate the stability issue, as the terms in the upper and lower bounds can be quite different for larger intervals.

We, therefore, refine the solution with tight upper and lower bounds:
\begin{equation}
\begin{split}
    &\text{softmax}(\iv{z}) = \left[ \frac{\exp(\ul{z}_c)}{\sum_{i \neq c}{\exp(\ol{z}_i)} + \exp(\ul{z}_c)}, \frac{\exp(\ol{z}_c)}{\sum_{i \neq c}{\exp(\ul{z}_i)} + \exp(\ol{z}_c)} \right]\\
    &= \left[ \frac{\exp(\ul{z}_c - a_l)}{\sum_{i \neq c}{\exp(\ol{z}_i - a_l)} + \exp(\ul{z}_c - a_l)}, \frac{\exp(\ol{z}_c - a_u)}{\sum_{i \neq c}{\exp(\ul{z}_i - a_u)} + \exp(\ol{z}_c - a_u)} \right]\\
    & a_l = \max\{\ul{z}_c, \ol{z}_{i \neq c}\},\  a_u = \max\{\ol{z}_c, \ul{z}_{i \neq c}\}.
\end{split}
\end{equation}

We prove soundness through monotonicity. Softmax is monotonically increasing in $z_c$:
\begin{equation}
    \text{softmax}(z) = \frac{\exp(z_c)}{\sum_{i=1}^m{\exp(z_i)}} = \frac{\exp(z_c)}{\exp(z_c) + b} = 1 - \frac{b}{b + \exp(z_c)}, b \in \R_{\geq 0},
\end{equation}
and monotonically decreasing in $z_{i \neq c}$:
\begin{equation}
    \text{softmax}(z) = \frac{\exp(z_c)}{\sum_{i=1}^m{\exp(z_i)}} = \frac{b_1}{\exp(z_i) + b_2}, \{b_1, b_2\} \in \R_{\geq 0}.
\end{equation}
Therefore, using \cref{eq:monotonic-increasing} and \cref{eq:monotonic-decreasing}, our bounds are sound and tight.

The same concept can be extended to the logsoftmax function:
\begin{equation}
\begin{split}
    &\text{logsoftmax}(\iv{z}) \\&= \Biggl[
        \ul{z}_c - \log\Biggl(\sum_{i \neq c}^m{\exp(\ol{z}_i)} + \exp(\ul{z}_c) \Biggr),
        \ol{z}_c - \log\Biggl(\sum_{i \neq c}^m{\exp(\ul{z}_i)} + \exp(\ol{z}_c) \Biggr)
    \Biggr]\\
     &= \Biggl[
        (\ul{z}_c - a_l) - \log\Biggl(\sum_{i \neq c}^m{\exp(\ol{z}_i - a_l)} + \exp(\ul{z}_c - a_l) \Biggr),\\
        & \qquad (\ol{z}_c - a_u) - \log\Biggl(\sum_{i \neq c}^m{\exp(\ul{z}_i - a_u)} + \exp(\ol{z}_c - a_u) \Biggr)
    \Biggr]\\
    & a_l = \max\{\ul{z}_c, \ol{z}_{i \neq c}\},\  a_u = \max\{\ol{z}_c, \ul{z}_{i \neq c}\}
\end{split}
\end{equation}
The $\log$ function is monotonically increasing.
That implies that $\log \circ \text{softmax}$ has the same monotonicity as $\text{softmax}$, which proves both the soundness and tightness of our bounds.

\subsection{Binary Cross-Entropy}
\label{app:relaxations:bce}

For binary classification problems, we use the binary cross-entropy (BCE) loss:
\begin{equation}
    J(p, y) = -(y \log(p) + (1 - y) \log (1 - p),\ p = \sigma(z) = \frac{1}{1 + \exp(-z)}.
\end{equation}

Combined with the sigmoid activation of the last layer, it has the same numeric stability issue as cross-entropy with softmax.
The typical solution looks, therefore, similar:
\begin{equation}
\begin{split}
    J(z, y) &= - (y \log(\sigma(z)) + (1 - y) \log (1 - \sigma(z))\\
    &= z - zy + \log(1 + \exp(-z))\\
    &= z - zy + a + \log(\exp(-a) + \exp(-z -a))\\
    &a = \max(-z, 0).
\end{split}
\end{equation}

BCE is monotone in $z$. Therefore, we can directly define tight and sound bounds using \cref{eq:monotonic-increasing}:
\begin{equation}
    J(\iv{z}, y) = \left[ \min\{L(\ul{z}, y), L(\ol{z}, y)\}, \max\{L(\ul{z}, y), L(\ol{z}, y)\} \right]
\end{equation}

For the backwards pass, we get the derivative of BCE as
\begin{equation}
    \frac{\partial J}{\partial z}(z) = \sigma(z) - y
\end{equation}

This function is, again, monotonically increasing in $z$. Therefore
\begin{equation}
    \frac{\partial J}{\partial z}(\iv{z}) = [\sigma(\ul{z}) - y, \sigma(\ol{z}) - y]
\end{equation}
are tight and sound bounds.

\section{Additional Experiments}
\label{app:experiments}

We supplement our main experiments in \cref{sec:experiments} with additional experiments to analyze the influence of different aspects of model training on \method.

\subsection{Complexity and Compute}

Training with interval bounds has the same asymptotic complexity as regular model training.
The overhead due to computing upper and lower bounds for each intermediate result is a constant factor.

In practice, the overhead is larger, as our BoundFlow library is not as heavily optimized as modern machine learning libraries like PyTorch.
For practical comparison, training a fully connected network with two layers for 100 epochs on \twomoons\ with \method{} takes 7.6 seconds, while the same model without bounds in PyTorch trains in 0.5 seconds.
For \mnist, a single epoch with \method{} takes 2 seconds, while the same model without bounds in PyTorch trains in 0.38 seconds per epoch.

We run all experiments on a Linux machine with an 8-core 3.6 GHz CPU, 32GB of RAM, and an Nvidia Titan RTX GPU.
For exact software versions, please refer to the environment file provided with the implementation.

\subsection{Training Details}

The experiments in \cref{sec:experiments} use the following hyper-parameters as defaults:
\begin{itemize}
    \item batch size: 100
    \item learning-rate: 0.01 Two-Moons, 0.05 MNIST 1/7
    \item max-epochs: 100
    \item architecture: 3-layer MLP with ReLU activation
\end{itemize}

For \twomoons, we use a training set size of 1000 samples, and 200 samples for validation and test sets, respectively.
For \mnist, we use the standard train and test splits of the full MNIST dataset, filtered to only include samples from classes 1 and 7.
We use 2000 images from the training set for validation, which leaves us with 9831 samples for training and 1938 samples for the test set.

\subsection{Influence of Hyper-Parameters}
\label{app:experiments-hyperparameters}

In this set of experiments, we evaluate the influence of different training parameters on \method.
To this end, we train fully connected models on the \twomoons dataset without pertaining while varying different hyper-parameters.

\begin{table}[ht]
    \centering
    \setlength{\tabcolsep}{6pt}
    \begin{tabular}{ccccc}
    \toprule
    $\epsilon$ & 0.0001 & 0.001 & 0.01 & 0.1 \\
    \midrule
    certified accuracy  & 83.9 $\pm$ 3.6 & 82.2 $\pm$ 4.4 & 71.5 $\pm$ 11.2 & 36.7 $\pm$ 14.8 \\
    \bottomrule
    \end{tabular}
    \vspace{1em}
    \caption{
        Certified accuracy of 2-layer MLPs trained on \twomoons{} for different $\epsilon$.
        All values show mean and standard deviation over 10 independent training runs with different, randomly chosen seeds.
        For smaller perturbations budgets, the algorithm consistently converges toward an optimum.
        For large perturbations, some initializations cause training to converge due to one of the effects analyzed in \cref{sec:convergence}, leading to lower mean accuracy and higher variance.
    }
    \vspace{-1em}
    \label{tbl:epsilon}
\end{table}

The most influential hyper-parameter is the perturbation radius $\epsilon$.
$\epsilon$ is defined by the threat model, and it directly controls the adversary's capabilities.
The larger $\epsilon$, the stronger the attacker's potential influence on the prediction.
\Cref{tbl:epsilon} shows certified accuracies for 2-layer fully-connected models with ReLU activation with different perturbation budgets.
The certified accuracy is averaged over 10 runs with different, randomly chosen seeds for initialization.
For small $\epsilon$, \method{} consistently converges towards an optimum with high certified accuracy.
This changes for large $\epsilon$, where some runs fail to converge depending on initialization.
This is likely due to one of the effects discussed in \cref{sec:convergence} and results in a lower average certified accuracy and higher standard deviation.

\begin{table}[ht]
    \centering
    \setlength{\tabcolsep}{6pt}
    \begin{tabular}{lccccc}
        \toprule
        learning rate & 0.01 & 0.1 & 1.0 & 5.0 & 10.0 \\
        \midrule
        certified acc. & 71\% & 71\% & 75\% & 74\% & 72\% \\
        \bottomrule
    \end{tabular}
    \vspace{1em}
    \caption{
        Certified accuracy of 2-layer MLPs trained on \twomoons\ with $\epsilon=0.01$ and different learning rates.
        The experiments show convergence and good certified accuracy across four orders of magnitude.
    }
    \label{tbl:learning-rate}
    \vspace{-1em}
\end{table}

\Cref{tbl:learning-rate} shows certified accuracy when trained with different learning rates.
Training is stable across four magnitudes of learning rates with certified accuracies consistently above 70\%.
The increase in certified accuracy with higher training rates is likely due to the much faster convergence towards the optimum, which decreases the number of learning steps and therefore the number of accumulated over-approximations.

\begin{table}[ht]
    \setlength{\tabcolsep}{6pt}
    \centering
    \begin{tabular}{lcccc}
        \toprule
        batch size & 50 & 100 & 500 & 1000 \\
        \midrule
        certified accuracy & 72.5\% & 72.5\% & 70.5\% & 70.5\% \\
        \bottomrule
    \end{tabular}
    \vspace{1em}
    \caption{
        Certified accuracy of 2-layer MLPs trained on \twomoons\ with $\epsilon=0.01$ and different batch sizes.
        The results show that certified accuracy is independent of batch size.
    }
    \label{tbl:batch-size}
    \vspace{-1em}
\end{table}

\Cref{tbl:batch-size} shows certified accuracy when training with different batch sizes.
The results are nearly identical, demonstrating that \method{} works independently of batch size.

\begin{table}[ht]
    \setlength{\tabcolsep}{6pt}
    \centering
    \begin{tabular}{lc}
        \toprule
        hidden nodes & certified accuracy \\
        \midrule
        10 hidden nodes per layer & 71.5\% \\
        20 hidden nodes per layer & 72.5\% \\
        30 hidden nodes per layer & 79.0\% \\
        40 hidden nodes per layer & 63.0\% \\
        \bottomrule
    \end{tabular}
    \vspace{1em}
    \caption{
        Certified accuracy for MLPs with different layer sizes trained on \twomoons{} with $\epsilon = 0.01$.
    }
    \label{tbl:model-architecture}
    \vspace{-1em}
\end{table}

Lastly, we show that \method{} works with different numbers of hidden connections.
\Cref{tbl:model-architecture} shows certified accuracy for MLPs with different node counts.
In general, the larger the model, the higher certified accuracy up to a point, where the effects of over-approximation increase with increasing model size, and the bounds become less precise as a result.

\end{document}